\documentclass[10pt,twocolumn,letterpaper]{article}

\usepackage{cvpr}
\usepackage{times}
\usepackage{epsfig}
\usepackage{graphicx}
\usepackage{amsmath}
\usepackage{amssymb}
\usepackage{array} 

\usepackage{color}

\usepackage{changes}

\usepackage{authblk}

\setremarkmarkup{\color{red} [#2]}

\usepackage[pagebackref=true,breaklinks=true,letterpaper=true,colorlinks,bookmarks=false]{hyperref}

\cvprfinalcopy 

\def\cvprPaperID{****} 

\ifcvprfinal\fi
\begin{document}

\title{Weakly Supervised Attention Learning for Textual Phrases Grounding }


\author[1]{Zhiyuan Fang}
\author[2]{Shu Kong}
\author[1]{Tianshu Yu}
\author[1]{Yezhou Yang}
\affil[1]{\{zy.fang, tianshuy, yz.yang\}@asu.edu \ \ \ \ \ \ \ \ Arizona State University, Tempe, USA}
\affil[2]{skong2@ics.uci.edu \ \ \ \ \ \ \ \  \ \ \ \ \ \  \ \ \ \ \ \  \ \ \ \ \ \ \ \ \ \ \ \ \ \ University of California, Irvine, USA}

\maketitle

\begin{abstract}

Grounding textual phrases in visual content is a meaningful yet challenging problem with various potential applications such as image-text inference or text-driven multimedia interaction. Most of the current existing methods adopt the supervised learning mechanism which requires ground-truth at pixel level during training. However, fine-grained level ground-truth annotation is quite time-consuming and severely narrows the scope for more general applications. In this extended abstract, we explore methods to localize flexibly image regions from the top-down signal (in a form of one-hot label or natural languages) with a weakly supervised attention learning mechanism. In our model, two types of modules are utilized: a backbone module for visual feature capturing, and an attentive module generating maps based on regularized bilinear pooling. We construct the model in an end-to-end fashion which is trained by encouraging the spatial attentive map to shift and focus on the region that consists of the best matched visual features with the top-down signal. We demonstrate the preliminary yet promising results on a testbed that is synthesized with multi-label MNIST data.
\end{abstract}

\section{Introduction}

We study the problem of textual phrases grounding for images in the context of computer vision and natural language processing. Recent works like image-text query \cite{rui1999image} are designed to retrieve a set of images which have the largest similarities considering both visual aspects and textual descriptions. Going beyond instances querying, language (textual) grounding emphasizes the capability of localizing a word or a phrase in a given image. Most of the prior works in this domain have benefited significantly from cognitive performance improvements especially the object detection, object classification and semantic
segmentation. More specifically, previously works like \cite{yu2018mattnet,chen2017query} consider the grounding problem into two steps: the proposals generation and the attribute matching. During testing phase, the proposals/regions with the highest matching probability with the textual descriptions would be selected as the target region. 
Though the  strategy of using bounding boxes produces the location of the objects much more precisely, it largely depends on the object annotations and is accompanied with heavy computational redundancy on regions feature extraction. Moreover, the supervised learning paradigm is quite domain specific and it fails to achieve satisfactory results when we do not have sufficient labeled data for the a new task to train a reliable grounding model \cite{yeh2018unsupervised}. Thus, facing the necessity of using bounding box annotations under such scenario, we need a more general method.

\begin{figure}[t]
\begin{center}
\includegraphics[width=1.0\linewidth]{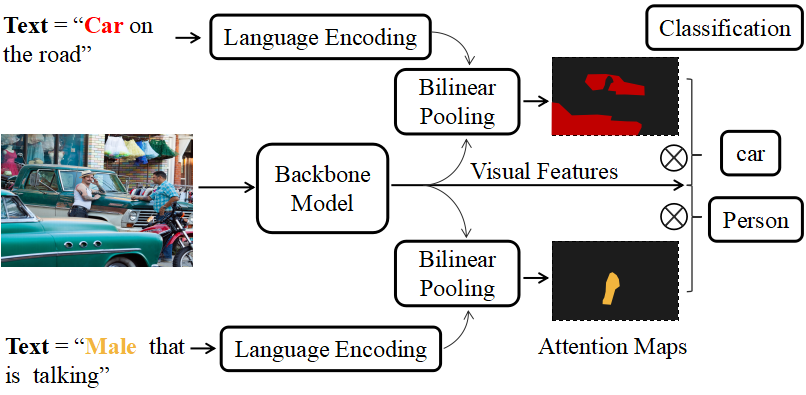}
\end{center}
\vspace{-2mm}
\caption{Illustrative framework of textual phrases grounding: We propose to localize phrases in images by matching  visual features with phrases. For example, from the phrase “man that is talking”, we can force the attentive map to focus on the region contains the target objects by classify the highlighted feature region as "person" and "talking".}
\vspace{-2mm}
\label{fig:abstract}
\end{figure}

\begin{figure*}
\begin{center}
\includegraphics[width=1.0\linewidth]{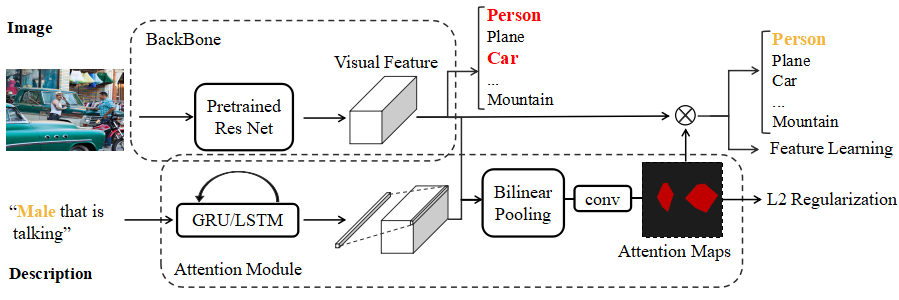}
\end{center}
\vspace{-2mm}
\caption{The overall architecture of our textual grounding pipeline consists of two part: one backbone module for global visual feature extraction, and a bilinear pooling based attentive module for attentive map generation. After word embedding, the expanded textual vector together with the visual features will be sent to  bilinear pooling layer, based on which the attentive map is generated. The training of our networks adopts an end-to-end manner with on a joint supervision signal: a cross-entropy loss for classification and a pair regularizing loss for the feature supervision.  In addition, a l2 regularization loss is also applied on the attentive map for sparsity. }
\vspace{-2mm}
\label{fig:flow}
\end{figure*}

We here put forward a weakly supervised textual grounding mechanism based on an attention generating method \cite{fukui2016multimodal}. Our hypothesis is that the deep convolutional features would still be spatial consistent with the input image\cite{hou2017deep}, and representations from a well trained deep model could be scale\cite{wang2017multi} and viewpoint invariant on the same objects \cite{gong2014multi}. Similar to the object co-segmentation \cite{joulin2010discriminative} which typically refers to the task of jointly segmenting ``something similar'' in a given set of images, we encourage the attentive map to search out to the regions which contain the similar features in a given set of images (see Fig.~\ref{fig:abstract}). In the same time, similar to the work in \cite{wen2016discriminative,zhang2016range}, a feature learning loss is applied on the architecture to obtain the deep features with more inter-class dispersion and intra-class compactness.

We evaluate our approach on a synthesized multi-MNIST dataset obtaining testing samples as the ones visualized in Fig.~\ref{fig:example1}. In this extended abstract, we demonstrate that our proposed pipeline is applicable for supervised/unsupervised demands and provides more interpretability and flexibility when transferring to different domains that is lack of fine-grained annotations, and it shows promising avenue to further conduct a comprehensive set of quantitative experiments.







\section{Methods}
We formulate here the weakly supervised grounding task as selecting the best region $r$ in an image  $I$ given a set of objects $\{O_1, O_2 ... , O_N\}^N_{i=1}$. More specifically, the task selects the best matched feature region from the feature maps. Given a pretrained backbone model $F$, we first extract global visual feature maps $f = F(I)$, and an attentive map/region $r = A(f, O_i)$, where $A$ refers to a bilinear pooling based attentive module. A classifier $C$ classifies $O_i$ based on the global visual features after applying attentive maps to it: $C(f\bigotimes r) = O_i$ (see Fig. \ref{fig:flow}). Besides, for different images $I_i$ and $I_j$ that contain a common object $O$, their region features are enforced to be similar: $f_i\bigotimes A(f_i, O) \approx f_j\bigotimes A(f_j, O)$ and this is achieved by the feature learning loss. 

\subsection{Visual Feature Extraction}
In order to extract global and discriminative features from the image, we adopt the deep pretrained classification model as the backbone to extract the deep features. It is a pre-trained deep model, e.g., Residual Network pretrained on ImageNet, which consists of about 15 million high-resolution labeled images in roughly 22 000 categories. The model has been shown that the encapsulated representations contained within can work remarkably well for a large set of diverse image classification tasks and often outperforms other classification approaches. In our two experiments, we use a well trained residual-18 net on MNIST as the backbone module for the multi-number grounding task to perform as the feature extractor. To maintain the resolution of the attentive map, here we remove all the max pooling layers in the original architecture and take the output from the last convolutional layer as the visual features. 

\subsection{Concatenation of Multi-Modal Feature}
Since our system aims to utilize the given top-down signals (either in the form of word embedding or one hot vector) to guide the attentive module, we need an appropriate feature fusion/concatenation method. Traditional methods  such as the  element-wise product, sum, or the concatenation of the visual and textual representations, are not as expressive as an outer product of the visual and textual vectors. Therefore, we turn to the bilinear pooling here. Bilinear pooling is designed to fuse the multi-modal features by outer product and has been proved to be much more expressive than the above mentioned methods in a series of tasks \cite{kong2017low}. However, direct applying the outer product is typically infeasible due to the curse of dimensionality, the  Multimodal Compact Bilinear pooling (MCB) \cite{gao2016compact} works nicely for our schema to maintain the outer product feature with lower dimensions. Concretely, the Count Sketch projection function \cite{charikar2002finding} $\Psi$ would be applied on the outer product of the visual feature $f$ and expanded top down signal $t$ for dimensionality reduction: $f^{'}=\Psi (f)\ast \Psi(t)$. If converted to frequency domain the concatenated outer product can be written as: $f^{'}= {FFT}^{-1}(FFT(\Psi (f))\bigodot FFT(\Psi(t)))$. Based on the $f^{'}$, which consists of multiplicative interaction between all elements of both vectors, attentive map $r$ is further computed after several convolutional layers. 

\subsection{Weakly Supervised Training}
To shift the attentive map to the targeted region, the training of our networks yields an end-to-end manner with on a joint supervision signal: a cross-entropy loss $L_{ce}$ for classification and a pair regularizing loss $L_p$ for the feature supervision. In addition to the joint signal, an $L_{2}$ regularization $R$ is applied on the attentive map to ensure sparsity. Thus, we formulate the overall loss $L$ as:
\begin{equation} \label{eq1}
\begin{split}
L & = L_{ce} + \lambda \cdot{L_p} + \beta \cdot R ,
\end{split}
\end{equation}
where $L_{ce}$ refers to the cross-entropy loss, and $L_p$ can be expressed as:
\begin{equation} \label{eq1}
\begin{split}
L_2 & = \begin{cases}
||f_i^{'} - f_j^{'} - m||^2,& \text{if } y_i = y_j\\
-||f_i^{'} - f_j^{'}||^2.             & \text{otherwise}
\end{cases}
\end{split}
\end{equation}

The $f^{'}$ represents the attentive high-lightened features, $y_i$, $y_j$ are class labels that are most relevant/matched to the input phrases in the images $I_i, I_j$. And, 
\begin{equation} \label{eq1}
\begin{split}
R & = \frac{1}{n^2}\sum_{x,y=0}^{n}{r_{(x,y)}^2},
\end{split}
\end{equation}
where  $n$ denotes the size of the the attentive map $r$, and $x$, $y$ denote the coordinates of each pixel. With an ablation study, we show that a regularization term can reduce and eliminate the invalid attention noises effectively, thus making the attentive map even more compact and sparse.   
Practically, our designed networks that are supervised by joint loss is trainable and can be optimized by standard stochastic gradient decent (SGD) methods. Scalar $\lambda$ and $\beta$ are two hyper parameters controlling the balance among the three loss functions, and the margin $m$ is the minimum margin distance among the intra-class features. The conventional cross entropy loss can be considered as a special case of this joint supervision, if $\lambda$ and $\beta$ is set to be $0$.

\section{A Preliminary Experiment}
In this Section, we present a preliminary experiment on a synthesized multi-MNIST dataset to validate the feasibility of our attentive mechanism. To better simulate a real textual grounding scenario, a multi-MNIST dataset is introduced. While instead of ground natural expressions/phrases, we feed in one-hot class labels as our top-down signal, and our goal is to localize all the matched regions in that image that contain the target numbers without using any bounding boxes annotation.

\begin{figure}[h!]
\begin{center}
\includegraphics[width=1.0\linewidth]{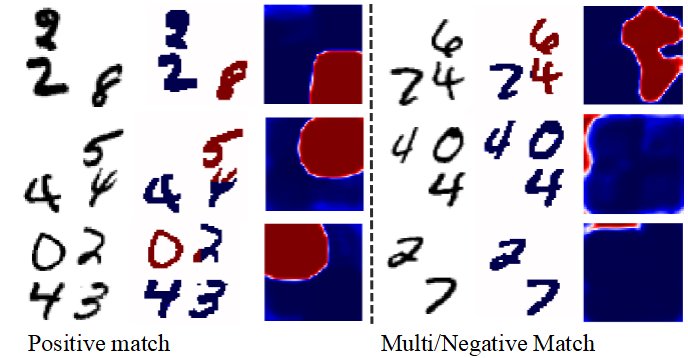}
\end{center}
\vspace{-2mm}
   \caption{Example of localization in multi MNIST (best viewed in color). Starting from left to right, we present the input images, followed by the grounded region (grounded regions are colored in red), and the attentive map. The input textual signals here are \lq8\rq, \lq5\rq, and \lq0\rq, respectively. For the right part of the gallery, we show the grounding result with multiple input signals (\lq6\rq, and \lq4\rq, in this sample) in the upper row, the attentive map when input signal doesn't exist (\lq2\rq, in this sample, middle row), and a failure case in the lower-left row.}
\label{fig:example1}
\end{figure}

\textbf{Data Preparing:} We generate a multi-MNIST dataset in which each image contains randomly $1$ to $4$ in number of $0$-$9$ handwritten digits (see Fig. \ref{fig:example1} as an example). In order to make the synthetic data more challenging, we randomly increase the overlapping ratio among the numbers and add Gaussian white noise at the same time. 500k images are generated in total together with their corresponding labels and annotations. 

\textbf{Network Structure:} ResNet-18 serves as our backbone architecture and we extract features from the final convolutional layers of the $4$-th stage as the visual features . We modify the stride in all convolutional layers to be $1$ to maintain feature resolution which is the same as the input image.  As for the attentive module, since we adopt one-hot encodings in this setting, we replace the GRU/LSTM unit with a 128 units dense connected layer. We then expand the textual input to be with the same size of visual features, and pass the two tensors through MCB.  After MCB pooling, two $1\times 1$ convolutional layers are applied to generate a one channel attentive map. 

\textbf{Training Process:} We pretrain our backbone architecture on the original MNIST dataset for classification with an accuracy of 99.40\%. After that, at the second stage of end-to-end training, we set the initial learning rate of the whole model to be $0.1$ and further reduce the learning rate for the pretrained backbone to be always $1/10$. We also test joint training and make the backbone architecture trained from scratch, and as expected, the former training schema converges much faster. We believe this is due to that a well pretrained model provides a comparatively precise local feature that enables the feature matching to converge more quickly. During the training procedure, we set two hyper-parameters $\lambda$ and $\beta$ to be $0.1$, respectively.

\textbf{Result:} We show quantitatively the effectiveness of our approach in Table~\ref{table:generaresult1} and \ref{table:generaresult2}, We adopt the standard evaluation metric for both the detection and the localization tasks (a.k.a. the mean average precision): if the IoU with the ground truth box is bigger than a certain threshold, that localization is considered to be correctly identified. In Table~\ref{table:generaresult1} that among different settings of the $\lambda$, our model can achieve up to $67.04\%$ IoU for the localization task. It is worth noting that the attentive map are evaluated directly without any further post-processing steps. We further compare the 
average IoU achieved under the two training schema: training from scratch v.s.  fine-tuning. Though we observe that using a well trained backbone model as the feature extractor is able to accelerate the training phase to a great extend, the final performances achieved are comparable, indicating that an end-to-end training from scratch is also applicable for our approach.
In Table~\ref{table:generaresult2} we further report the mean average precision value under three IoU thresholds. In practice,  we consider the prediction is correct if the IoU with ground truth is higher than $0.5$, in which case our model achieved $71.53\%$ mAP. 

To further illustrate our model's outputs, we show some of the grounding results in Fig.~\ref{fig:example1}. Our approach is able to capture the correct object described by the input query (the left row in Fig. \ref{fig:example1}). When the input query doesn't exist in the image, an almost blank attentive map would be generated as expected (the second row in the right column in Fig. \ref{fig:example1}). We also show a failure case (the lower-left one).

\begin{table}
\small
\begin{center}
\begin{tabular}{lrrrr}
\firsthline
\multicolumn{5}{c}{ Avg IoU } \\
\cline{2-5}
Model        &$\beta$@0       &$\beta$@0.05    &$\beta$@0.1      & $\beta$@0.2 \\
\hline\hline
Fine-tune backbone     &59.74    &60.84   &\textbf{67.07} &36.37\\
Train from scratch       &57.90    & 63.39   &\textbf{66.40} &41.02   \\
\lasthline
\end{tabular}
\end{center}
\vspace{-2mm}
\caption{The intersection over union (IoU) values of the different training schema on multi-MNIST datasets with our proposed architecture. Fine-tune backbone model refers to an end-to-end architecture with a pre-trained single handwritten number recognition model. Performances using various scales for regularization are also reported.}
\vspace{-2mm}
\label{table:generaresult1}
\end{table}

\begin{table}
\small
\begin{center}
\begin{tabular}{l|rrr||r}
\firsthline
Group    &IoU@0.3       &IoU@0.4     &IoU@0.5       &Avg mAP \\
\hline\hline
3$\times$3    &\textbf{83.40}    &82.54   &60.03    &75.23   \\
2$\times$2   &94.48    &\textbf{96.22}   &71.53    &87.41   \\
\lasthline
\end{tabular}
\end{center}
\vspace{-2mm}
\caption{The mean average of precision achieved on testing datasets. Here 2$\times$2 denotes that each testing image consists of 4 numbers (Fig.~\ref{fig:example1}), and 3$\times$3 denotes the samples with $9$ numbers. The mAP is obtained over different thresholds applied.}
\vspace{-2mm}
\label{table:generaresult2}
\end{table}

\section{Conclusion}
In this extended abstract, we present a novel approach for weakly supervised textual phrases grounding based on regularized bilinear pooling and feature learning. We report a preliminary validation of our approach on a synthesized multi-label MNIST dataset. The results obtained in our preliminary experiments demonstrate a promising avenue to achieve expandability and feasibility by the weakly supervised training mechanism in the textual phrase grounding task. We are currently extending the architecture to conduct further model validation on a large benchmarking natural images dataset such as MSCOCO \cite{lin2014microsoft}.

{\footnotesize 
\bibliographystyle{ieee}
\bibliography{egbib}
}

\end{document}